\documentclass[runningheads]{llncs}

\usepackage{eccv}
\usepackage{graphicx}
\usepackage{booktabs}
\usepackage{multirow}
\usepackage{pifont}
\usepackage{textcomp}
\usepackage[accsupp]{axessibility} 
\usepackage[pagebackref,breaklinks,colorlinks,citecolor=eccvblue]{hyperref}
\usepackage{hyperref}

\begin{document}

\newsavebox\CBox
\def\textBF#1{\sbox\CBox{#1}\resizebox{\wd\CBox}{\ht\CBox}{\textbf{#1}}}
\title{Superior and Pragmatic Talking Face Generation with Teacher-Student Framework} 

\titlerunning{Abbreviated paper title}

\author{Chao Liang\inst{1} \and
Jianwen Jiang\inst{1} \and
Tianyun Zhong\inst{1} \and
Gaojie Lin\inst{1} \and
Zhengkun Rong\inst{1} \and
Jiaqi Yang\inst{1} \and
Yongming Zhu\inst{1}}

\authorrunning{Liang  et al.}
\institute{ByteDance Inc.}
\maketitle

\begin{figure}[h]
\centering
\includegraphics[width=1.00\linewidth]{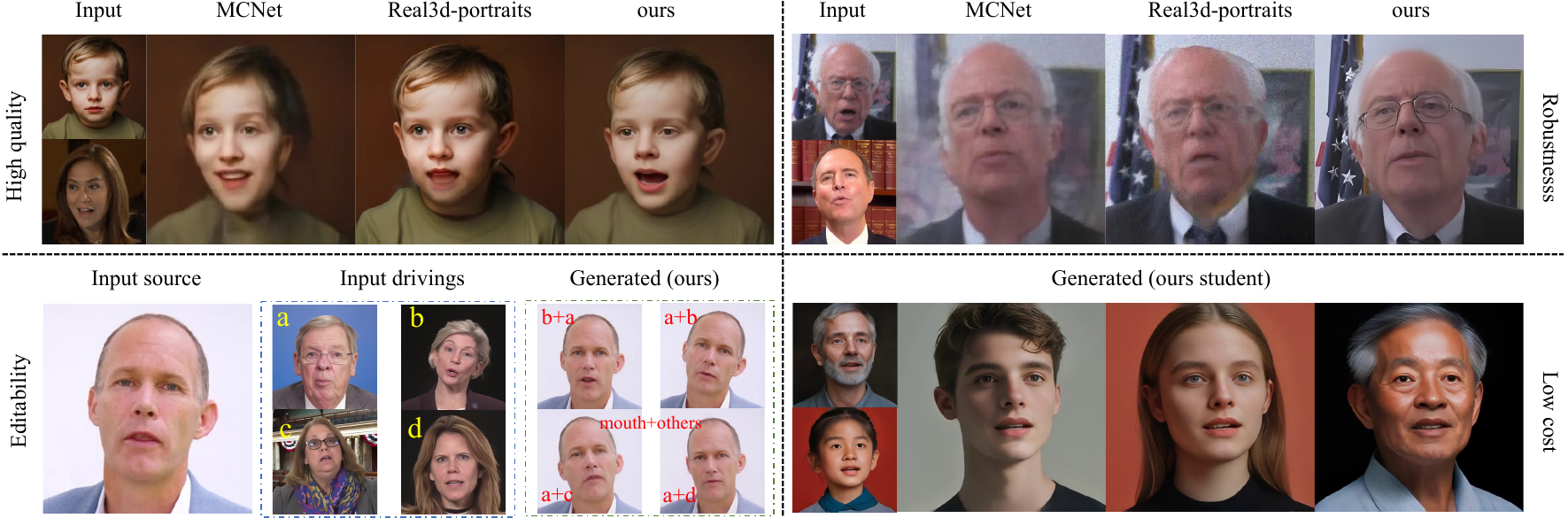}
\caption{Our proposed method outperforms state-of-the-art ones in the following aspects: 1) \textbf{Quality}: producing higher-quality results (top left); 2) \textbf{Robustness}: maintaining robustness even with poor-quality input (top right); 3) \textbf{Editability}: enabling users freely edit facial attributes (bottom left); 4) \textbf{Low cost}: achieving comparable result by distillation with 99\% reduction in FLOPs (bottom right).}\label{fig1}

\end{figure}
\begin{abstract}

Talking face generation technology creates talking videos from arbitrary appearance and motion signal, with the "arbitrary" offering ease of use but also introducing challenges in practical applications. Existing methods work well with standard inputs but suffer serious performance degradation with intricate real-world ones. Moreover, efficiency is also an important concern in deployment. To comprehensively address these issues, we introduce \textbf{SuperFace}, a teacher-student framework that balances \textBF{quality}, \textBF{robustness}, \textBF{cost} and \textBF{editability}. We first propose a simple but effective teacher model capable of handling inputs of varying qualities to generate high-quality results. Building on this, we devise an efficient distillation strategy to acquire an identity-specific student model that maintains quality with significantly reduced computational load. 
Our experiments validate that SuperFace offers a more comprehensive solution than existing methods for the four mentioned objectives, especially in reducing FLOPs by 99\% with the student model.
SuperFace can be driven by both video and audio and allows for localized facial attributes editing. 
\href{https://superfacelink.github.io/}{Project Page.}

\end{abstract}

\section{Introduction}
\label{sec:intro}

Talking face synthesis has been applied in various scenarios, and related technologies such as video translation and online conferences have significantly impacted our daily lives. In recent years, the field has witnessed marked advancements. However, prevailing attempts frequently concentrate on boosting performance with regular inputs, overlooking the complex factors encountered in practical applications. Specifically, users might provide inputs with inconsistent quality, such as blurry appearance or intense motion, which is extremely challenging to handle. Besides, current methods have not adequately addressed the issues of cost-effectiveness and editability, both of which are significant concerns for real-world deployment. In general, an excellent method is expected to be:

\begin{itemize}
\item[$\bullet$]

\textit{High-quality.} 
Quality encompasses motion accuracy, visual clarity, as well as the robustness to handle extreme inputs. 
Despite considerable progress, previous works fail to fulfill both simultaneously. Many works \cite{prajwal2020lip, siarohin2019first, wang2021one, Zhou2021Pose, ren2021pirenderer} made substantial efforts on precise motion control, yet due to their coarse motion representations, they still fall short in handling large movements and subtle facial expressions. 
Recent works \cite{hong2023implicit, Zhang_2023_CVPR, zhong2023identity, ye2024real3d} achieve better visual quality under standard inputs through ID-aware techniques, but stably generating clear results remains elusive for the diverse and variable inputs in the real world. 

\item[$\bullet$]

\textit{Pragmatic.} The efficiency of algorithms shapes their practicality, especially when high quality has already been achieved. Unfortunately, state-of-the-art methods like MCNet \cite{hong2023implicit} and Real-3d \cite{ye2024real3d} always demand extensive computing consumption for over 600G FLOPs to produce high-resolution $512 \times 512$ outputs, which imposes great costs in real-world applications. Due to the trade-off between computational expense and quality, directly reducing it tends to degrade the results, so it is essential to develop a novel framework that excels in both high quality and low cost. Moreover, processing cross-modal inputs and editing local facial attributes are also common user-expected capabilities, which existing methods often struggle with.

\end{itemize}

To comprehensively tackle the practical challenges mentioned above, We propose SuperFace, a teacher-student based framework that delivers a \textbf{high-quality}, \textbf{robust}, \textbf{low-cost}, and \textbf{editable} solution for talking head generation, as shown in Fig. \ref{fig1}. We first designed a potent teacher model to ensure a high-quality and robust generation. At its core stands: 1) a simulation for super-resolution (SSR) strategy that trains the model's generation and robustness capabilities using constructed low-quality and high-quality pairs end-to-end, enabling the teacher to create high-quality results from varied-quality features. 2) a motion-enhancing mechanism (MEM) that leverages 3D convolutional networks and efficiently integrates 3D facial priors via multi-stage infusions, enabling the model to refine precise motion depiction.

Equipped with the powerful teacher model, we turned our attention to cost. To break the trade-off between performance and efficiency, we introduce a distillation strategy to train an identity-specific student model that compresses the computational load by 99\%. During distillation, the teacher's components and intermediate features can be delivered to the student without any cost, serving either as specific modules or inputs. With just one image for two hours' training, the student model achieves performance comparable to that of the teacher, with the ability of real-time inference on more low-end devices. In our experiments, we verified that the student model requires two orders of magnitude fewer FLOPs than current state-of-the-art methods.

To further enable flexible editing of facial attributes, we introduce a simple mask training mechanism (MTM) that masks and replaces the global driving signals generated by MEM with other local ones. 
Based on this, an auxiliary audio-to-lip module is proposed to provide specific control for lip movements, promoting speech-driven settings. These expanded functionalities offer the benefits of crossmodal-driven settings and local editing, allowing for greater versatility in practical synthesis.

In summary, our technical contributions are as follows:
\begin{itemize}
\item[$\bullet$] We propose SuperFace, a teacher-student framework, which avoids the zero-sum trade-off between generation quality and computational cost and achieves superior and pragmatic talking head generation.

\item[$\bullet$] We propose SSR to learn image quality enhancement and MEM to facilitate accurate motion both end-to-end, in the teacher model. Experiments demonstrate that SuperFace not only significantly outperforms existing methods in quality and robustness but also possesses crossmodal-driving and facial attribute editing abilities.

\item[$\bullet$] Based on teacher model, we propose a distillation scheme to train an identity-specific student model that achieves performance similar to that of the teacher while reducing computational costs by two orders of magnitude.

\end{itemize}

\section{Related Work}
\label{sec2}

Existing researches \cite{siarohin2019first, ren2021pirenderer, wang2021one, Zhao_2022_CVPR, Zhang_2023_CVPR} in this field primarily concentrate on enhancing visual quality and motion replication performance.

For visual quality, some works \cite{mallyaimplicit, zhang2023dinet, zhong2023identity} extract ample appearance information from a set of source images to minimize occluded regions. Nonetheless, usability challenges are introduced due to the harsh condition of multiple images. MCNet \cite{hong2023implicit} learns a global facial representation space to compensate more correlated information using the appearance pool, and overcomes the ambiguous generation caused by the dramatic motions.
Inspired by the StyleGAN series \cite{karras2019style, karras2020analyzing, karras2021alias}, some works \cite{yin2022styleheat, burkov2020neural, sun2022ide} borrow their ideas or utilize their pre-trained models to regenerate clear talking head video. However, they also exhibit poor controllability inherited from StyleGAN. Some works \cite{ye2024real3d, Zhang_2023_CVPR} incorporate pre-trained super-resolution networks as post-processing modules to improve image quality, at the expense of disrupting the end-to-end structure and increasing the inference cost. Another disadvantage of these schemes is that they neglect time consistency, resulting in an temporal incoherent generated video.

For motion accuracy, many studies \cite{siarohin2019first, zhang2023sadtalker, mallyaimplicit, wang2021one} symbolize motion as a representation consisting of a set of learned keypoints along with their dense fields. Such implicit semantics bring challenges for downstream tasks such as local editing. To simplify the training process, some works \cite{kim2018deep,kim2019neural, fried2019text, zhou2020makelttalk, ren2021pirenderer, Zhang_2023_CVPR} utilize explicit 3D Morphable Models (3DMMs) \cite{blanz1999morphable,paysan20093d} or 2D facial landmarks \cite{kowalski2017deep, guo2019pfld} directly, which significantly increase the interpretability of the model. However, this simplistic usage fails to explore the full potential of these facial priors in describing motion, and may introduce additional errors from the detection tools.

Besides the GAN-based approaches mentioned above, there are also some NeRF-based \cite{guo2021ad, Ma_2023_CVPR, ye2024real3d} and Diffusion Model-based \cite{Shen_2023_CVPR, Mukhopadhyay_2024_WACV, he2023gaia, du2023dae} efforts. AD-NeRF \cite{guo2021ad} is the first NeRF-based method, it generates a dynamic neural radiance field using audio and renders it to a talking video. Real-3d \cite{ye2024real3d} employs a large image-to-plane model to distill 3D prior knowledge for better visual quality. Despite these innovations, they hardly produce photo-realistic results due to the inherent difficulty in modeling high-frequency details with volume rendering. The Denoising Diffusion Probabilistic Models \cite{ho2020denoising} perform better than GANs in some image generation tasks, and thus emerges many studies in this field. Some works \cite{he2023gaia, du2023dae} utilize diffusion model to predict motion latents and render the final video via VAE. EMO \cite{tian2024emo} successfully applied the diffusion model to rendering, bypassing the need for intermediate 3D models or facial landmarks. The biggest drawback of these methods is their substantial computational cost, making deployment difficult.

To enable the application of deep networks with large parameter counts on lower-end devices, researchers have not only designed lightweight structures \cite{howard2017mobilenets, zhang2018shufflenet} but have also proposed techniques such as parameter pruning \cite{lecun1989optimal, wen2016learning}, parameter quantization \cite{courbariaux2015binaryconnect, gong2014compressing}, and knowledge distillation \cite{buciluǎ2006model}. Unfortunately, this line of research still remains largely unexplored in talking head generation.

\section{Methodology}

The goal of this work is to achieve \textBF{high-quality}, \textBF{robust}, \textBF{low-cost}, and \textBF{editable} generation simultaneously. We address these issues progressively using a teacher-student framework. For convenience, we define the teacher model as $\Phi$ and the student model as $\phi$. The organizations of this section are as follows: We introduce the superb teacher model in Sec. \ref{sec3d1}. In Sec. \ref{sec3d2}, we describe the details of our distillation paradigm and student model. In Sec. \ref{sec3d3}, we explain how to expand the functionalities including crossmodal-driven settings and local facial attribute editing. Finally, in Sec. \ref{sec3d4}, we introduce the training details as well as the loss functions for both models.

\subsection{Teacher Model}
\label{sec3d1}

\begin{figure}[h]
\centering
\includegraphics[width=1.00\linewidth]{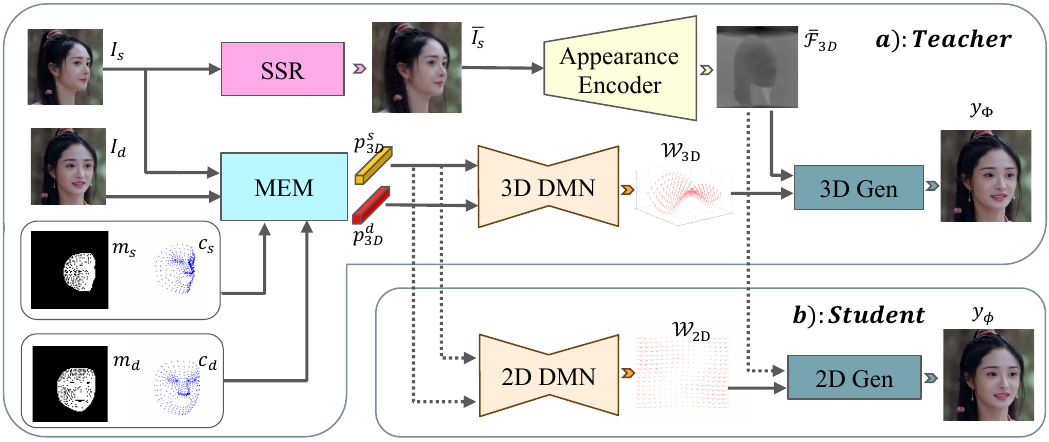}
\caption{The pipeline of our proposed framework. SuperFace consists of a teacher model and a student model. a) We first train an extremely powerful teacher model using MEM (motion-enhancing mechanism) and SSR (simulation for super-resolution). b) Then we distill its knowledge into an efficient student through feature delivery.}
\label{fig2}
\end{figure}

\noindent \textbf{Standard procedure.} Talking Head Generation aims to animate a static portrait using driving signals. We start with its standard procedure in a video-driven setting. For easy understanding, readers may refer to the top of Fig. \ref{fig2} while disregarding the SSR and MEM modules. Given two images $I_s\in\mathbb{R}^{3\times H\times W}$ refers to the source image and $I_d\in\mathbb{R}^{3\times H\times W}$ refers to the driving one, the teacher model first extracts a spatial appearance feature $\mathcal{F}$ from $I_s$ using an appearance encoder. After the motion field $\mathcal{W}$ is extracted from both $I_s$ and $I_d$ using a dense motion network (DMN), it will be used to warp $\mathcal{F}$, and the warped appearance feature $\mathcal{F}^{s \to d}$ will be fed into the generator (GEN) for decoding a realistic facial image.

\noindent \textbf{Motion-enhancing-mechanism}. Motion field $\mathcal{W}$ aims to warp the appearance features $\mathcal{F}$ of 
$I_s$ into $\mathcal{F}^{s \to d}$ to align with the motion status of $I_d$. In the real world, head motion is three-dimensional, thus $\mathcal{W}$ and $\mathcal{F}$ should contain depth information to be 3D-aware for better 3D-consistent feature alignment. Consequently, we process them into 3D spatial forms $\mathcal{W}_{3D}\in\mathbb{R}^{C\times{H^{\prime}}\times{W^{\prime}}\times{D^{\prime}}}$ and $\mathcal{F}_{3D}\in\mathbb{R}^{C\times{H^{\prime}}\times{W^{\prime}}\times{D^{\prime}}}$ with the help of 3D convolutional network. However, our experiments revealed that capturing precise 3D motion solely from planar pixel inputs $I$ is challenging for 3D networks, and additional 3D priors are necessary to provide more reliable information.

\begin{figure}[t]
\centering
\includegraphics[width=1.00\linewidth]{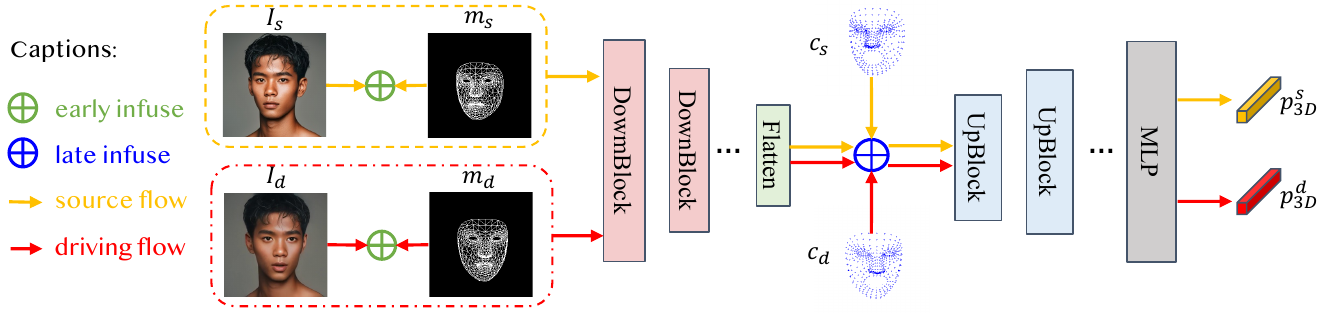}
\caption{Our proposed MEM incorporates 2D inputs and 3D priors to accurate depict of the motion. 3D priors are fully utilized through early/late infusion and two distinct representation designs.}

\label{fig3}
\end{figure}

With this in mind, we propose Motion-enhancing-mechanism (MEM) that leverages 3D face mesh to predict neural keypoints $p_{3D} \in ^{k \times 3}$, which denote the image's dynamic state and serve as the inputs to the DMN. 
There are two naive ways to infuse 3D face mesh: 1) treating it as a volumetric feature, which necessitates expensive computational costs; 2) converting it into a point cloud-like form, which sacrifices the inherent spatial relationships. To avoid these two drawbacks, we convert the 3D mesh into a dual representation: 1) a 2D mesh image that captures the specific topological relationships of the face mesh, and 2) a 3D landmarks that preserves the original facial depth information. This approach maintains the spatial and connectivity relationships of the face mesh with a lower computational cost. Furthermore, MEM develops a simple yet effective multi-stage infusion. As shown in Fig. \ref{fig3}, 2D mesh image $m$ and 3D landmarks $c$ are successively infused (early and late) to predict neutral keypoints $p_{3D}^{s/d}$, which serves as motion signals fed into DMN. The 2D image $I$ also serves as an input for MEM to model the motion of the background. Through multi-stage infusion of spatial and structural information from 3D facial mesh priors, MEM can assist the model in accurately capturing 3D-aware motion.

\begin{figure}[t]
    \centering
    \begin{minipage}[b]{0.49\linewidth}
        \includegraphics[width=\linewidth]{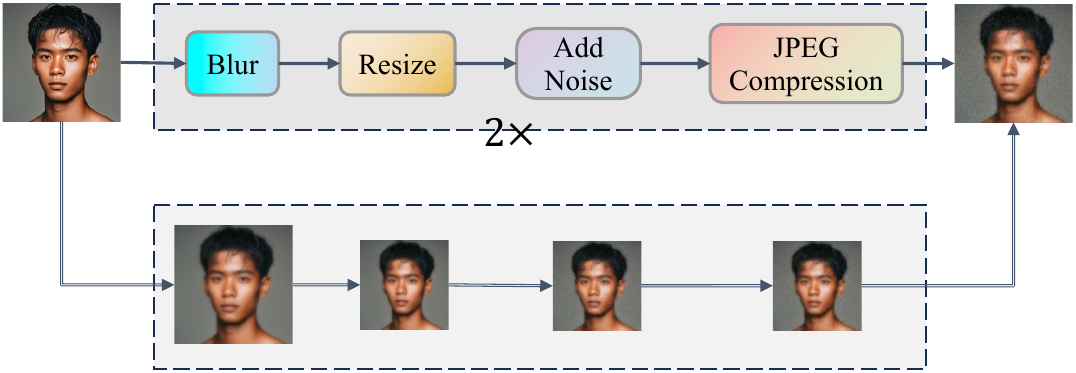}
        \caption{Pipeline of our SSR. We introduce a second-order degradation. Each order consists of random blurring, resizing, adding noise, and JPEG compression. }
        \label{fig4}
    \end{minipage}
    \hfill
    \begin{minipage}[b]{0.49\linewidth}
        \includegraphics[width=\linewidth]{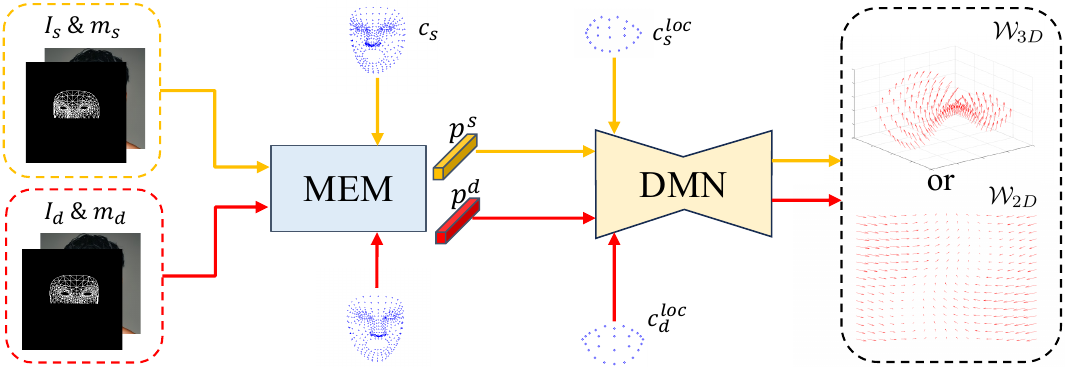}
        \caption{Pipeline of our MTM. We mask the vanilla driving signals and replace them with other ones. It facilicates local editing and crossmodel-driving. }
        \label{fig5}
    \end{minipage}
\end{figure}

\noindent \textbf{Simulation for super-resolution}. Generally, the quality of images generated by a model is consistent with its input, resulting in struggling when confronted with inferior priors. Although post-processing techniques can improve image clarity, there can be some side effects, as illustrated in Sec. \ref{sec2}. Inspired by works \cite{wang2021real, wang2021towards, yang2021gan, gu2022vqfr} in super-resolution, we introduce an end-to-end visual-enhancing training strategy without any post-modules. To be specific, a high-order degradation module is introduced to simulate real-world image degradation, as shown in Fig. \ref{fig4}, which is performed on source image $I_s$ to obtain the low-quality and high-quality image pairs $(\overline{I}_s, I_s)$. During training, $\overline{I}_s$ is used to extract the appearance features $\overline{\mathcal{F}}_{3D}$ for subsequent generation, while $I_s$ serves as the label for supervised learning. 
The process is similar to super-resolution tasks but can be implemented end-to-end within our model, hence we refer to it as \textBF{SSR} (Simulation for super-resolution). Thus, by incorporating SSR, the generator learns to robustly synthesize clear images stably from arbitrary features.

\subsection{Distillation Paradigm}
\label{sec3d2}
Utilizing MEM and SSR, we have developed a potent model equipped with the ability to model 3D-aware motion and ensure robust synthesis. Employing this as the teacher model, our goal is to lessen its computational demands, enabling us to derive a student model for practical deployment. To this end, we propose a distillation paradigm to distill teacher's knowledge into an identity-specific student model. Notably, the generalization of the student model and the expansion of identities are discussed in Sec. \ref{sec4d3}. 
As depicted in the bottom of Fig. \ref{fig2}, we design the student model with a structure similar to that of the teacher, yet significantly more computationally economical, replacing the 3D Convolutions with more lightweight 2D ones. During distillation, teacher model first infers portrait animations using a single image of a specific speaker and arbitrary driving videos. Both of the generated videos and intermediate features are then serve as training data of student model.
Thanks to its effective network design, the teacher model is capable of capturing precise motion details and modeling robust appearance features. To inherit these merits, the distillation paradigm deliver teacher's prior knowledge into student, including the motion features $p_{3D}$ for motion accuracy, and the appearance features $\overline{\mathcal{F}}_{3D}$ for visual quality.

Specifically, we employ an appearance distributor $\phi_d$ to compresse $\mathcal{W}_{3D}$ to $\mathcal{W}_{2D}$ and orthogonal projection to convert keypoints $p_{3D}$ to $p_{2D}$. Additionally, the teacher's discriminator is also employed to serve an effective supervisory role during the distillation process. Through such a design, the benefits of SSR and MEM proposed in Sec. \ref{sec3d1} remain highly advantageous for the student model. In experiments, we also verified that our proposed distillation strategy can significantly reduce the data and time dependence of distillation training.

Since the appearance and keypoint features are transmitted from the teacher model, the SSR and MEM proposed in our teacher model sitll work effectively within the student. Additionally, the teacher's discriminator is employed to serve an effective supervisory role during the distillation process.

\subsection{CrossModal and Local Editing}
\label{sec3d3}

So far, we have addressed the issues of high quality, robustness, and low cost, and now we turn our attention to the final challenge: editability.
Unfortunately, the ambiguous semantics of neural keypoints $p$ make it difficult to control the generation of local attributes. To address this, a semantically clearer and more easily learnable local driving signal is introduced. 

To eliminate the interaction between the different driving signals, we introduce a mask training mechanism (\textBF{MTM}). For convenience, we use the mouth landmarks $c^{loc}$ as an example of the driving signals that can be edited. As depicted in Fig. \ref{fig5}, by predicting the global driving signals $p$ from the masked inputs $I$ and $m$ and employing the local ones $c^{loc}$ as complement, MTM ensures that the semantics of two signals do not intertwine and guarantees the decoupling of motion control across different regions. During inferene, $c^{loc}$ and $p$ can originate from different sources to realize localized facial attribute editing. Notably, this design is generic and can be extended to arbitrary facial attributes.

Apart from extracting $c^{loc}$ directly from video, we also developed an additional audio2lip module to offer crossmodal-driving capabilities. This design provides the user with convenience that the control mechanism for localized editing is greatly simplified. 

\subsection{Training strategy}
\label{sec3d4}

To train SuperFace, we utilize loss functions similar to those used in previous works \cite{Zhao_2022_CVPR, wang2021one, siarohin2019first}. It comprises:

\noindent \textBF{Perceptual Loss $\mathcal{L}_{P}$.} The perceptual similarity loss \cite{johnson2016perceptual} between $I_d$ and $y$ helps to improve the perceptual quality. We minimize the pyramid perceptual loss between features extracted from $I_d$ and $y$ through a pre-trained network.

\noindent \textBF{GAN Loss $\mathcal{L}_{G}$.} . We employ the multi-scale PatchGAN \cite{wang2018high} as our discriminator, which predicts the probability whether the generated frame is comparable to the ground truth at the patch-level.

\noindent \textBF{Keypoint Loss $\mathcal{L}_{K}$.} To ensure that the extracted keypoints $p_{3D}$ evenly spread throughout the face region rather than crowd around a confined neighborhood, we incorporate a regularization as constraint. 
This involves measuring the distance between each point pair of $p_{3D}$, with the model incurring a penalty if any distance falls below a specified threshold.

\noindent \textBF{Expression Loss $\mathcal{L}_{Exp}$.} Experiments results indicate that the motion can't be accurately modeled  if the value of the expressions is too large. Therefore, we propose a regularization term to constrain its range.

\noindent \textBF{Head pose Loss $\mathcal{L}_{H}$.} A pre-trained network \cite{ruiz2018fine} is employed to detect the head poses of $y$ and $I_d$. We minimize $\mathcal{L}_{H}$ so that they are consistent.

\noindent \textBF{Equivariance Loss $\mathcal{L}_{Equ}$.} 
Due to $p_{3D}^c$ is learned unsupervised, we employ a equivariance loss to ensure its validity. 
Specifically, we apply a random transformation $T(\cdot)$ to $I_s$ and obtain the transformed image $I_{Ts}$. The keypoints $p_{3D}^c$ from $I_s$ and $p_{3D}^cT$ from $I_{Ts}$ should maintain the same transformation relationship.

\noindent \textBF{Reconstruction Loss $\mathcal{L}_{R}$.} This loss is to help the model converge quickly at the early stage of training and accelerate the training process.

\noindent \textBF{Local Loss $\mathcal{L}_{L}$.} To enhance the local high-frequency representation in the generated images, we crop the eyes and mouth regions and compute localized perceptual loss as well as landmarks loss.

For the teacher model, its total loss is:
\begin{equation}
\begin{aligned}
    \mathcal{L}_{\Phi} = &\lambda_{\Phi 1}\mathcal{L}_{P}
    +\lambda_{\Phi 2}\mathcal{L}_{G}
    +\lambda_{\Phi 3}\mathcal{L}_{K}
    +\lambda_{\Phi4}\mathcal{L}_{Exp} 
    +\lambda_{\Phi5}\mathcal{L}_{H} +\\
    &\lambda_{\Phi6}\mathcal{L}_{Equ}
    +\lambda_{\Phi7}\mathcal{L}_{R}
    + \lambda_{\Phi8}\mathcal{L}_{L}
\end{aligned}
\end{equation}

For the student model, its total loss is:
\begin{equation}
    \mathcal{L}_{\phi} =\lambda_{\phi 1} \mathcal{L}_{P}
    +\lambda_{\phi 2}\mathcal{L}_{G}
    +\lambda_{\phi 2}\mathcal{L}_{H}
    +\lambda_{\phi 4}\mathcal{L}_{R}
    +\lambda_{\phi 5} \mathcal{L}_{L}
\end{equation}

\section{Experiments}
\label{sec:blind}

In this section, we carry out comprehensive experiments to evaluate our SuperFace qualitatively and quantitatively. SuperFace consists of a powerful teacher model and a low-cost student model. We first introduce their specific implementations in Sec. \ref{4.1}. The evaluation of SuperFace's teacher model is then conducted in Sec. \ref{4.2}, spanning video-driven and audio-driven settings, as well as its robustness and editability. To verify the effectiveness of our distillation paradigm, Sec. \ref{sec4d3} presents the exploration results of the student model in various dimensions. Ablation studies in Sec. \ref{sec4d4} confirmed that each proposed component is essential and contributes to the ultimate performance.


\subsection{Implementation Details}
\label{4.1}
\subsubsection{Datasets}
Our teacher model was trained on the CelebV-HQ \cite{zhu2022celebv} dataset, a high-resolution video dataset with 35k videos at a minimum resolution of $512 \times 512$. A simple data filtering process was performed, and eventually, around $31,000$ videos were retained for use. The student model was subsequently trained using the teacher's inference results on the HDTF dataset \cite{zhang2021flow}. The test set for video-driven settings derives from three sources: TK1KH \cite{wang2021one}, HDTF \cite{zhang2021flow} and CCV2 \cite{hazirbas2021towards}, from which 300 videos were sampled randomly. 
Additionally, 100 HDTF videos entries with audio have been employed as our test set for audio-driven evaluation.

\subsubsection{Implementation Detail} We train our model on 8 NVIDIA A-100 (80G) machines. We used the ADAM
optimizer \cite{kinga2015method} with the learning rate $2\times10^{- 4}$ and $(\beta_1,\beta_2)=(0.5, 0.999)$. The numbers of keypoints $p$ fed into DMN is set to. The batch size is set to 32 for all experiments. To provide 3D priors and edit local facial attributes, we employ pretrained Mediapipe \cite{lugaresi2019mediapipe} to extract 3D landmarks and render meshes. As for the trade-off loss weights, we set $\lambda_{\Phi 1}=10, \lambda_{\Phi 2}=10, \lambda_{\Phi 3}=5, \lambda_{\Phi 4}=10, \lambda_{\Phi 5}=1,\lambda_{\Phi 6}=10,\lambda_{\Phi 7}=10,\lambda_{\Phi 8}=100, \lambda_{\phi 1}=10, \lambda_{\phi 2}=10, \lambda_{\phi 3}=10, \lambda_{\phi 4}=1, \lambda_{\phi 5}=100$.

\subsubsection{Metrics} Several metrics are utilized to assess our SuperFace’s validity in terms of both image quality and replication accuracy. For replication, we report the cosine similarity (\textbf{CSIM}) of face embedding between $I_s$ and $y$ using a recognition method \cite{deng2019arcface}. We also evaluated the whole motion transfer quality accessed by Average Keypoint Distance (\textbf{AKD}) \cite{chen2018lip}, Average Expression Distance (\textbf{AED}) \cite{radford2015unsupervised} and Head Pose Distance (\textbf{HPD}). For visual quality, since our SuperFace attempts to produce videos with higher clarity than the inputs, comparison-based metrics are not applicable in this context thus no-reference ones \textbf{Entropy} and \textbf{Energy} are employed. For audio-driven settings, we report SyncNet Distance (\textbf{Sync-D}) which measures the synchronization between audio-visual modalities. 

\subsection{Teacher Model Results} 
\label{4.2}

\begin{figure}[t]
\centering
\includegraphics[width=1.00\linewidth]{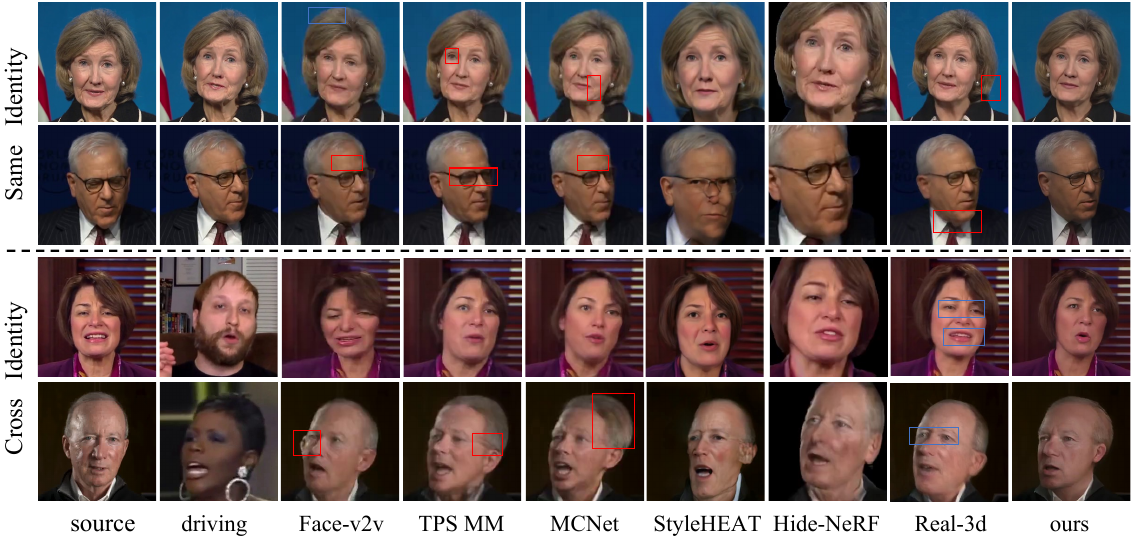}
\caption{Qualitative video-driven comparisons with baselines. For better visualization, we highlight the artifacts using different anchor boxes. The blue bounding box concentrates on motion accuracy, whereas the red one represents visual quality.}
\label{fig6}
\end{figure}

\begin{table}[tbh]
  \caption{Quantitative comparisons of our teacher with video-driven baselines.}
  \label{Tab1}
  \centering
  \begin{tabular}{ccccccccccc}
    \toprule
     \multirow{2}{*}{Methods} & \multicolumn{2}{c}{Visual} & \multicolumn{4}{c}{Same-id Motion} & \multicolumn{3}{c}{Cross-id Motion}
     & \multirow{2}{*}{\shortstack[c]{FLOPs \\ (G) $\downarrow$}} \\
     \cmidrule(l){2-3}
     \cmidrule(l){4-7}
     \cmidrule(l){8-10}
     &Ene$\uparrow$ & Ent/$10^6$ $\uparrow$ & CSIM$\uparrow$ & AKD$\downarrow$ & APD$\downarrow$ & AED$\downarrow$ & CSIM$\uparrow$ & APD$\downarrow$ & AED$\downarrow$ \\
    \midrule
    Face-v2v  & 4.80 & 7.02  &0.663 & 3.341& 1.294&0.101&0.468&4.459&0.197&600\\
    TPS MM & 4.81 & 6.34  & 0.668 & 1.897 & 1.195 & 0.083 & 0.416 & 4.435 & 0.152 & \textBF{120}\\
    MCNet & 4.76 & 5.07  & 0.668 & 1.570 & 1.322 & 0.086 & 0.378 & 4.413 & 0.154 & 200 \\
    ours 256 & \textBF{4.82} & \textBF{7.19} & \textBF{0.674} & \textBF{1.442} & \textBF{1.144} & \textBF{0.069} &\textBF{0.469} & \textBF{3.837}& \textBF{0.143} & 150 \\
    \midrule
    StyleHEAT & 4.80 & 3.95 & 0.337 & \texttimes{} & 2.417 & 0.148 & 0.277 & 4.778 & 0.200 & 1895\\
    Hide-NeRF & 4.29 & 5.41 & \texttimes{} & \texttimes{} & 2.721 & 0.155 & \texttimes{} & 4.962 & 0.210 & 1065 \\
    Real3d & 4.85 & 7.22  & 0.649 & 1.638 & \textBF{0.822} & 0.083 & 0.478 & 3.744 & 0.176 & 610\\
    ours 512 & \textBF{4.86} & \textBF{7.59} & \textBF{0.682} & \textBF{1.077} & 0.839 & \textBF{0.071} & \textBF{0.491} & \textBF{3.225} & \textBF{0.151} & \textBF{600} \\
  \bottomrule
  \end{tabular}
\end{table}

\subsubsection{Video-Driven}

We select several strong baselines using various technical routes. The quantitative results are presented in Tab. \ref{Tab1}. Obviously, our teacher model outperforms state-of-the-art methods across all objective metrics, expection of APD, where it is on the par with Real-3d. It is worth noting that our FLOPs are comparable to those of baselines when generating videos at the same resolution. We also present the qualitative comparisons in Fig. \ref{fig5}. It is evident that our approach surpasses these baselines in terms of generation quality and replication quality. Our technique is superior in capturing and reconstructing tiny movements like blinking, lip motion, etc. In addition, our SuperFace produces clearer portraits with rich texture details. High-frequency information such as gaze, hair, and wrinkles is successfully reconstructed.

Moreover, our teacher model shows great robustness in producing high-quality videos consistently, even when faced with large-posed or low-quality inputs. The corresponding qualitative results are available in Fig. \ref{fig6}. Since talking face synthesis is
a video generation task, we highly recommend the reader refer to our supplementary materials for more intuitive demos.

\begin{figure}[t]
\centering
\includegraphics[width=1.00\linewidth]{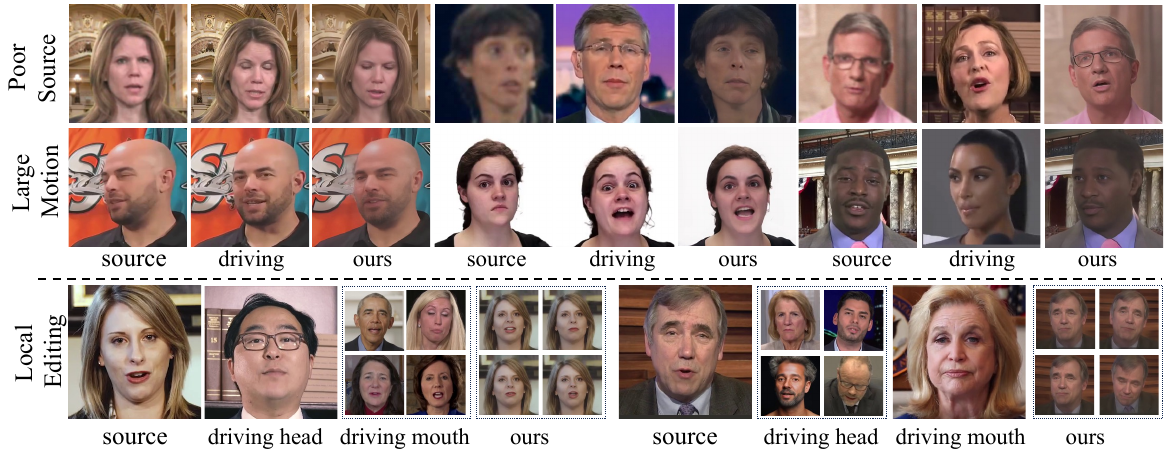}
\caption{More visualization results of our teacher model. Top : it produces high-quality video enven when faced with poor-quality appearance and intense-motion inputs. Bottom : it supports the editing of localized facial attributes.}
\label{fig6}
\end{figure}

\begin{table}[tb]
  \caption{Quantitative comparisons of our teacher with audio-driven baselines.
  }
  \label{Tab2}
  \centering
  \begin{tabular}{cccccc}
    \toprule
     Methods & AKD$\downarrow$ & APD$\downarrow$ & AED$\downarrow$ & CSIM$\uparrow$ & Sync-D$\downarrow$\\ 
    \midrule
    MakeItTalk  &7.245&1.616&0.127 & 0.711 & 9.460\\
    SadTalker & 7.299&1.003&0.115 & 0.678 & 7.867\\
    Real3d & 6.574&0.995&0.104 & 0.674 & \textBF{7.774}\\
    ours & \textBF{0.907}&\textBF{0.945}&\textBF{0.068} & \textBF{0.758} & 9.354\\
  \bottomrule
  \end{tabular}
\end{table}
\subsubsection{Audio-Driven} Audio-driven talking head generation places a greater emphasis on lip synchronization. Under this settings, we compare with MakeItTalk \cite{zhou2020makelttalk}, SadTalker \cite{zhang2023sadtalker} and Real3d-Protraits \cite{ye2024real3d}. The qualitative results are shown in Tab. \ref{Tab2}. It can be observed that our AKD metric substantially surpasses those of baselines, demonstrating SuperFace's attention to global facial attributes in audio-driven settings. 
In the realms of head motion and facial expressions, our teacher model has also realized the most accurate reconstruction. We also achieve better identity preservation performance.
As our trainging didn't incorporate the SyncNet which is used to calculate Sync-D, our teacher shows disadvantages compared to baselined that emolpyed it. Nevertheless, our performance is on par with other models.

\subsection{Student Model Results} 
\label{sec4d3}
\begin{figure*}[tb]
  \centering
  \subfloat[Steps-CSIM]{\includegraphics[width=0.45\textwidth]{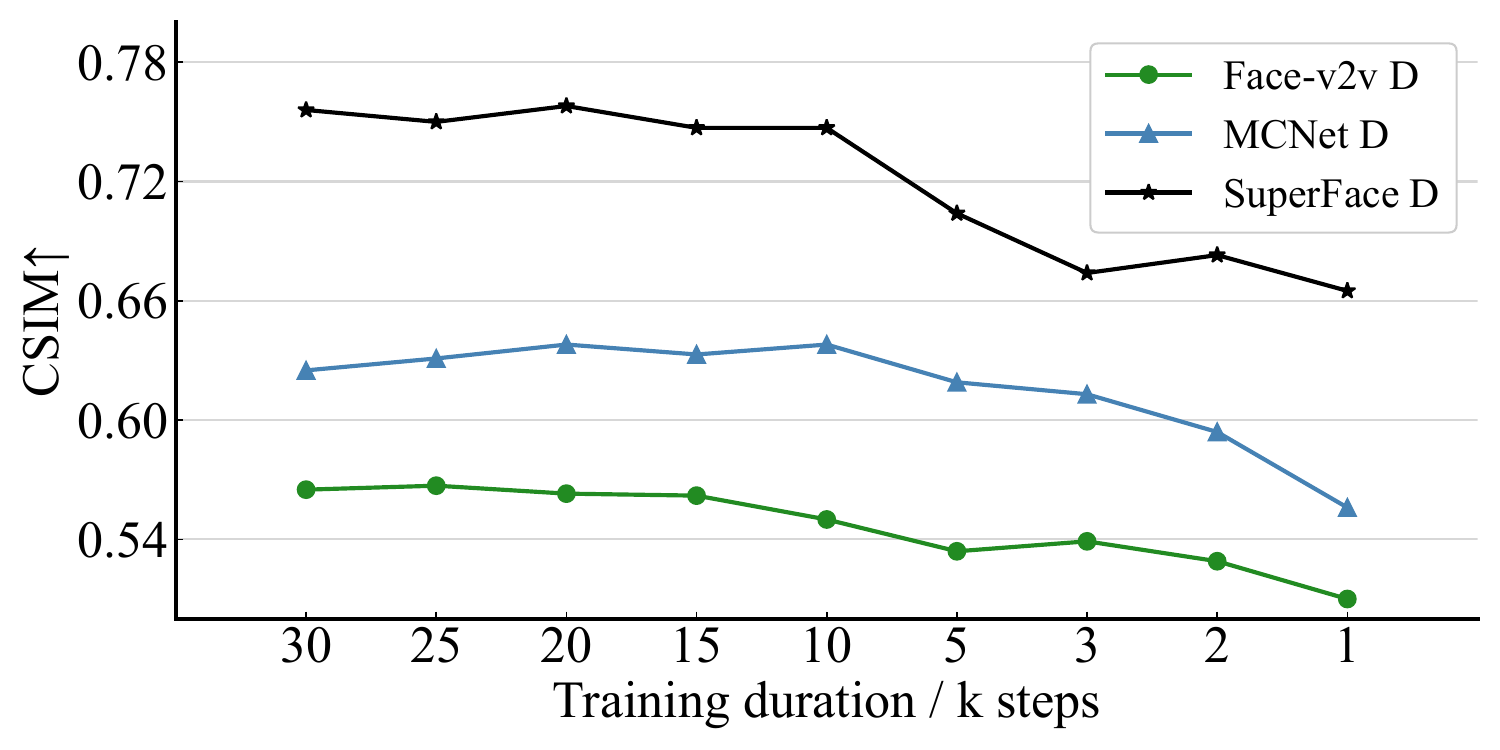}}
  \hfill 	
  \subfloat[Data-CSIM]{\includegraphics[width=0.26\textwidth]{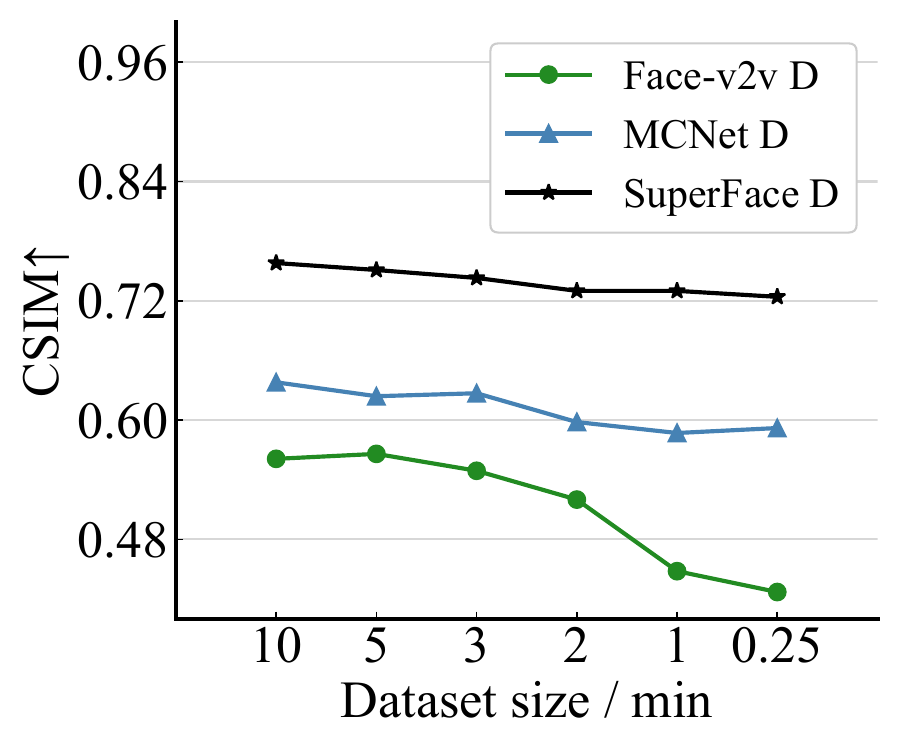}}
  \hfill 	
  \subfloat[FLOPs-CSIM]{\includegraphics[width=0.26\textwidth]{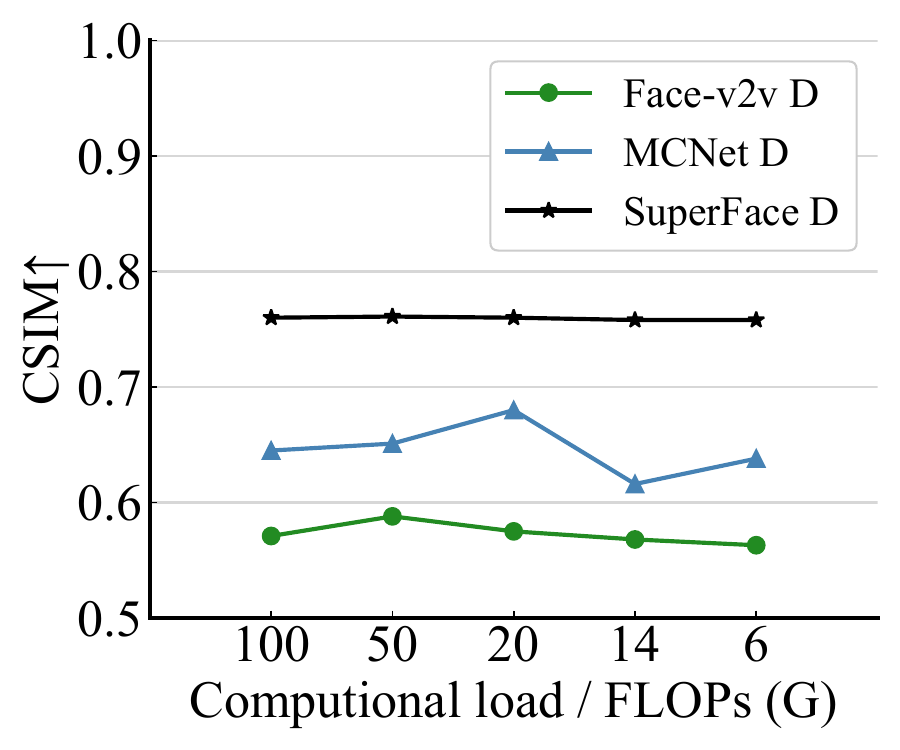}}
  \newline
   \subfloat[Steps-Energy]{\includegraphics[width=0.45\textwidth]{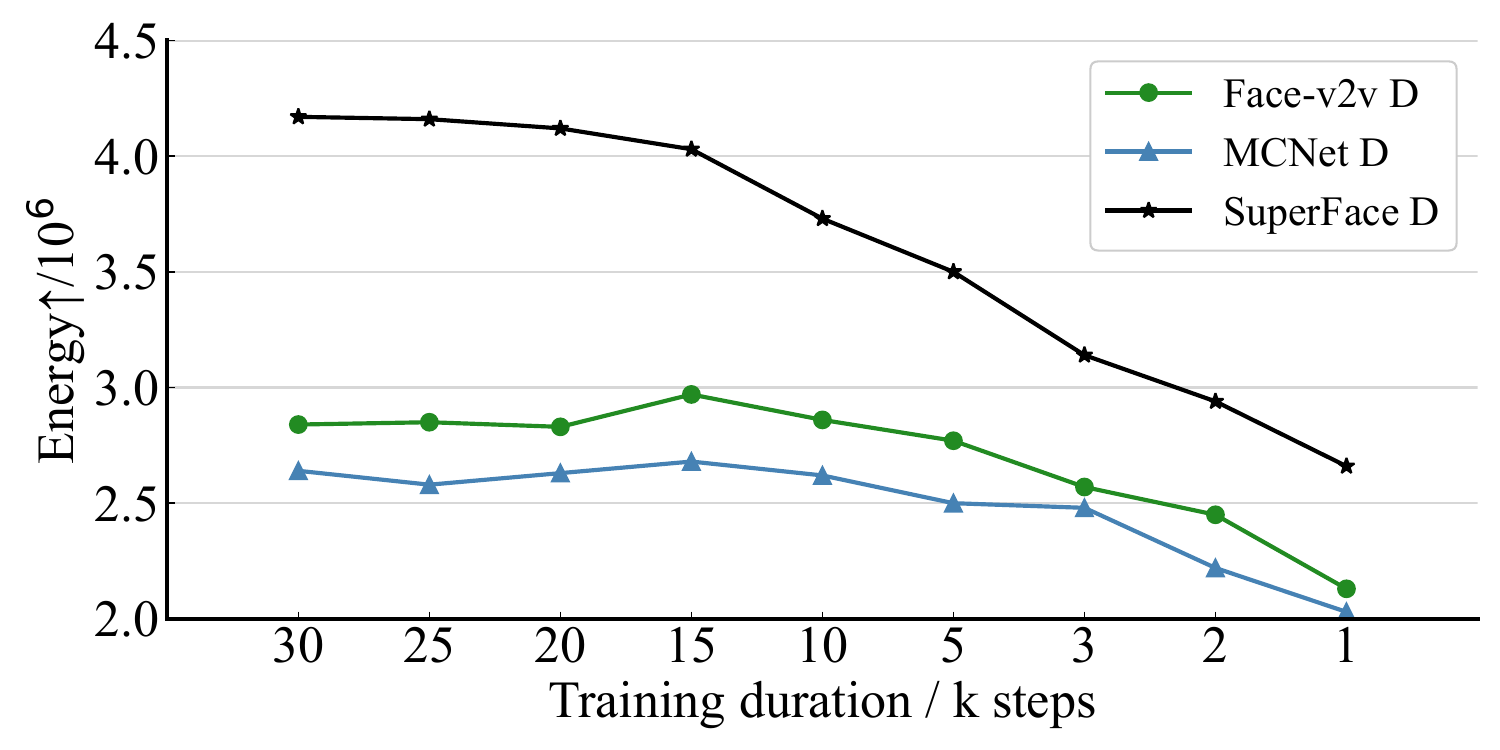}}
  \hfill 	
  \subfloat[Data-Energy]{\includegraphics[width=0.26\textwidth]{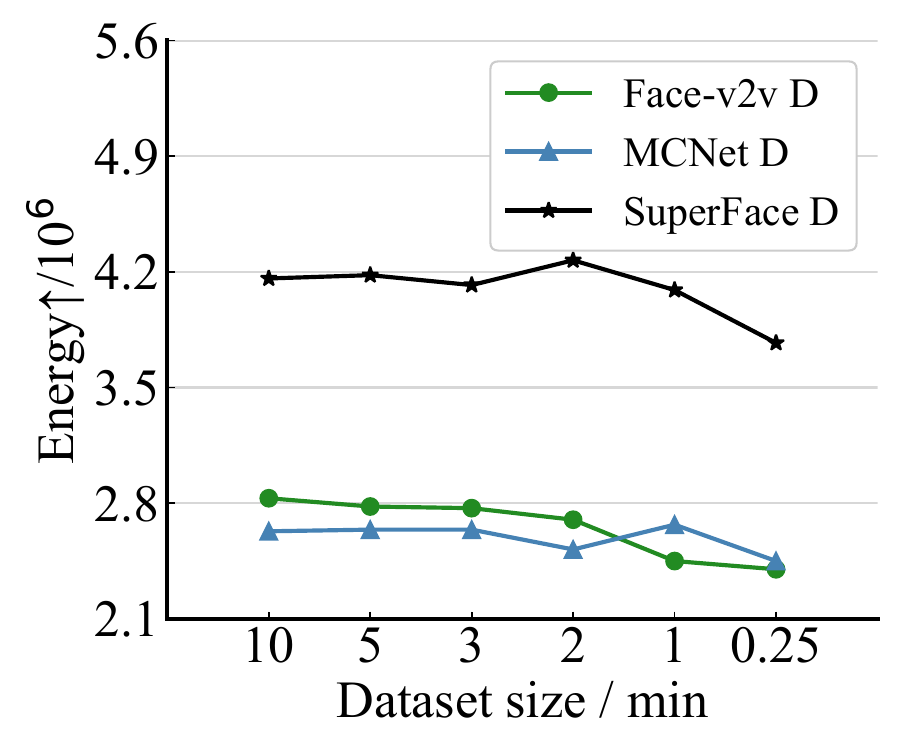}}
  \hfill 	
  \subfloat[Steps-Energy]{\includegraphics[width=0.26\textwidth]{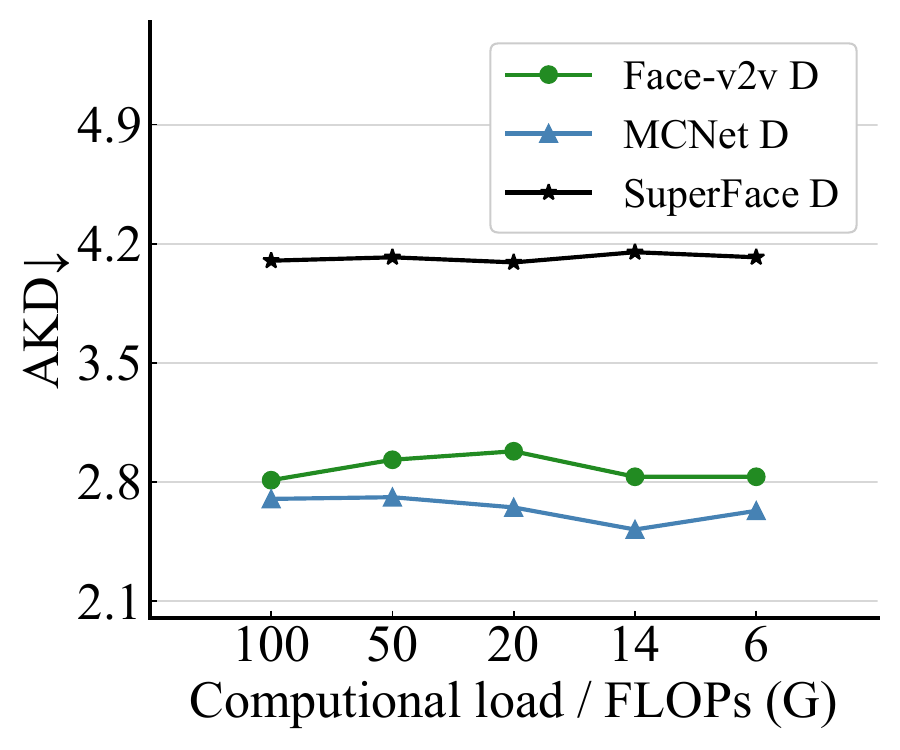}}
  \newline
  \subfloat[Steps-AKD]{\includegraphics[width=0.45\textwidth]{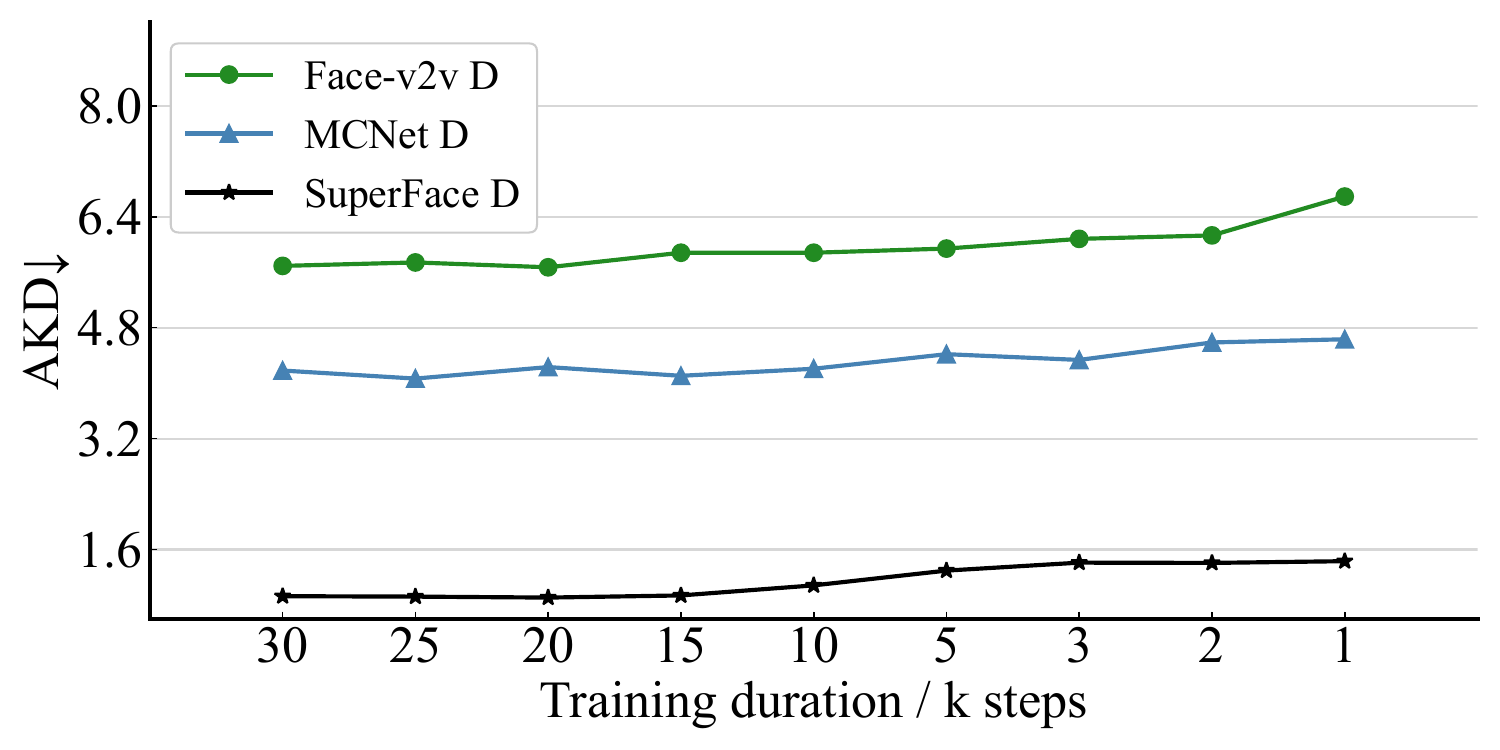}}
  \hfill 	
  \subfloat[Data-AKD]{\includegraphics[width=0.26\textwidth]{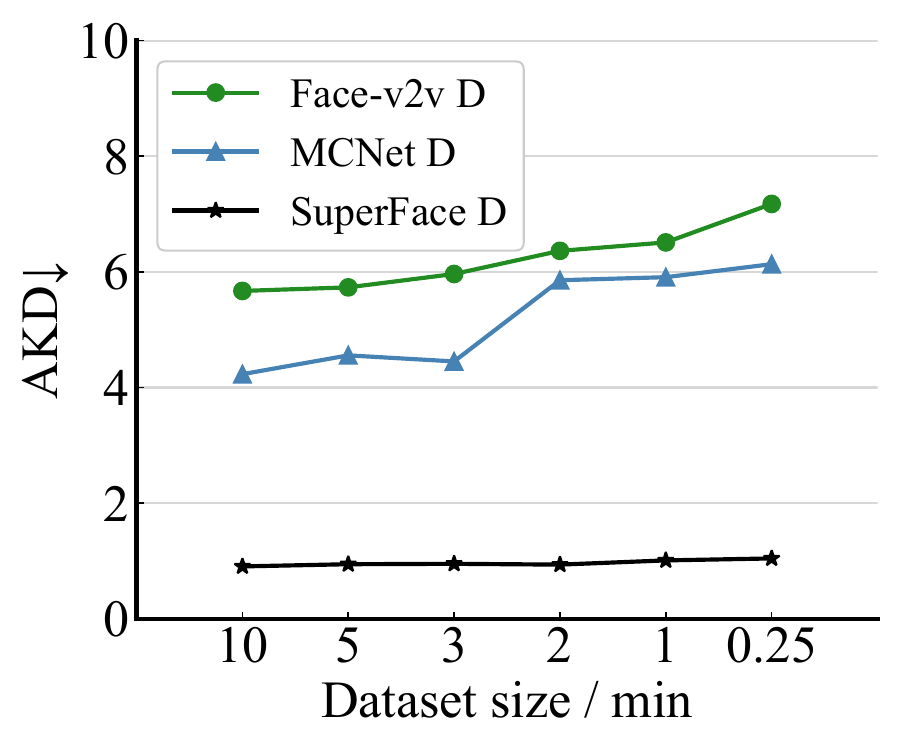}}
  \hfill 	
  \subfloat[FLOPs-AKD]{\includegraphics[width=0.26\textwidth]{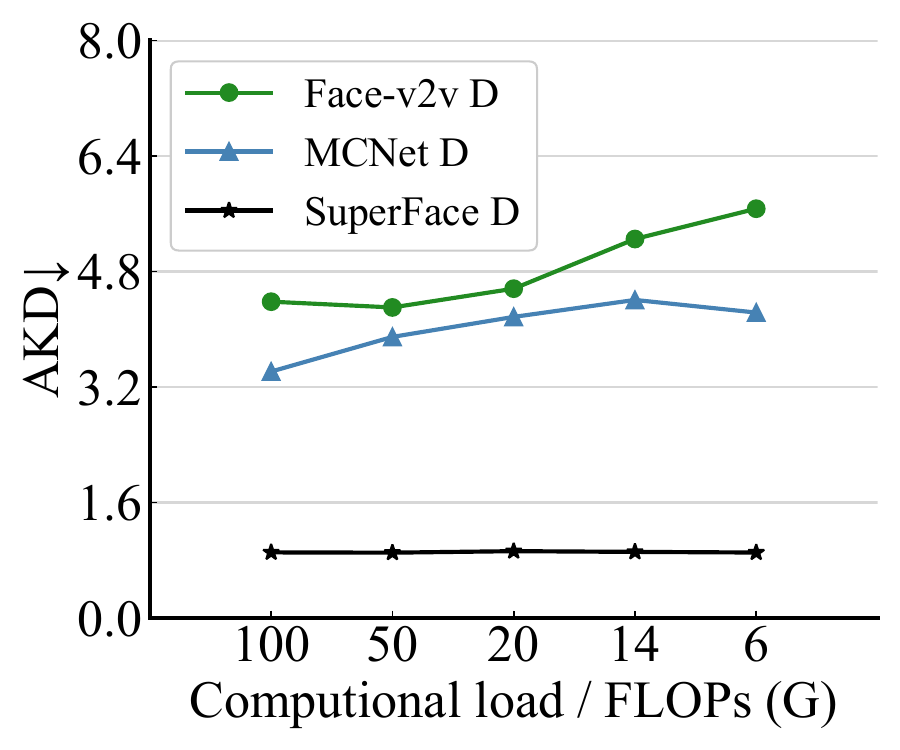}}
  \caption{The investigation results of the student model. From a row perspective, CSIM reflects the ability to preserve identity, Energy denotes visual quality, and AKD indicates the accuracy of the motion. Viewed column-wise, Steps represents an exploration into training duration, Data embodies an inquiry into the training datasize, and FLOPs pertains to the study of model complexity.}
  \label{Fig5}
\end{figure*}

\begin{table}[tb]
  \caption{Quantitative comparisons of our student model with baselines.
  }
  \label{Tab3}
  \centering
  \begin{tabular}{cccccccc}
    \toprule
     \multirow{2}{*}{Methods} & \multicolumn{2}{c}{Visual} & \multicolumn{4}{c}{Motion} 
     & \multirow{2}{*}{\shortstack[c]{FLOPs \\ (G) $\downarrow$}} \\
     \cmidrule(l){2-3}
     \cmidrule(l){4-7}
     &Ene$\uparrow$ & Ent/$10^6$$\uparrow$ & CSIM$\uparrow$ & APD$\downarrow$ & AKD$\downarrow$  & AED$\downarrow$ \\
    \midrule
    Face-v2v S & 4.41 & 2.26 &0.522 & 1.528& 6.643 & 0.248& 6.0\\
    MCNet S &  4.44 & 2.34  & 0.589 &1.625 & 5.237 & 0.188 & 6.0\\
    Face-v2v D &  4.51 & 2.83 & 0.563 & 1.373 & 5.678 & 0.174 & 6.0\\
    MCNet D &  4.57 & 2.63 & 0.638 & 1.319 & 4.231 & 0.143 & 6.0 \\
    ours & \textBF{4.78} & \textBF{4.16} & \textBF{0.758} & \textBF{0.399} & \textBF{0.907} &\textBF{0.069} & 6.0\\
  \bottomrule
  \end{tabular}
\end{table}

Given that early works in this field have overlooked computational costs, there are no baselines with comparable complexity to our student model. Based on teacher's baselines \cite{wang2021one, hong2023implicit}, we also established  baselines for the student by two principles. The first is simplifying them to 6G (consistent with our student model) while retaining their structure designs, then retrained their small models (S) using the same dataset and training strategy. The second is similar to our distillation paradigm, which learns knowledge from their large models to the distilled ones (D). The evaluation results are shown in Tab. \ref{Tab3}. It can be concluded that our student model has achieved a considerable lead in all metrics. Baselines, regardless of whether they are S or D, suffer severe performance degradation. Yet our student model manages to maintain consistency with the teacher model, and its visualized results in Fig. \ref{fig:8b} exhibit natural attributes and clear visual quality.
Another observation is that both S versions fall short of their D versions, suggesting that distillation is beneficial for reducing computational budgets.

As shown in Fig. \ref{Fig5}, we vary the data size, training steps, and FLOPs of the student model to explore its capabilities of identity-preserving (CSIM), visual quality (Energy), and motion accuracy (AKD). It is worth noting that our proposed model is not sensitive to dataset size or computational complexity (FLOPs), which only have a subtle influence on our model. Hence, it can be concluded that SuperFace's student model can quickly converge to a desirable state with minimal data (1 minute) and a short training duration (2$\sim$5 hours). Besides, our student model is scalable and behaves better when the training duration increases. 

\begin{figure}[tb]
  \centering
  \begin{subfigure}{0.49\linewidth}
   \includegraphics[width=1.00\textwidth]{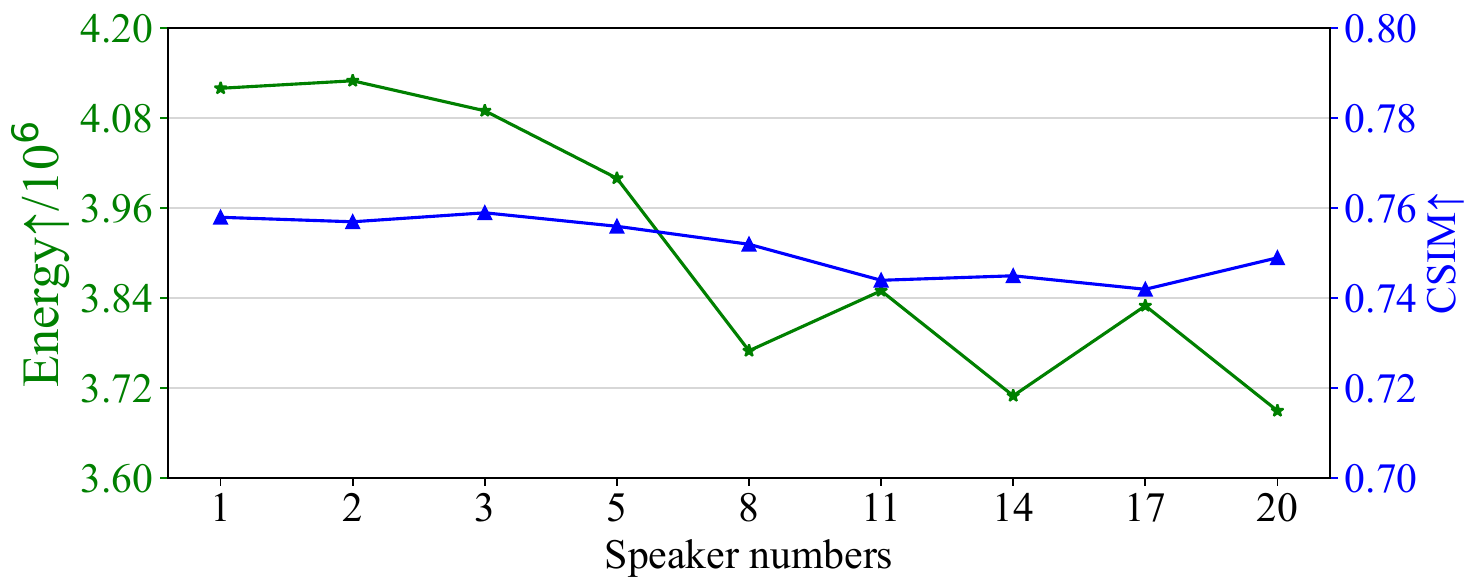}\label{fig:short-a}
    \caption{The curve of speaker numbers and model performance.}
    \label{fig:8a}
  \end{subfigure}
  \hfill
  \begin{subfigure}{0.49\linewidth}
    \includegraphics[width=1.00\textwidth]{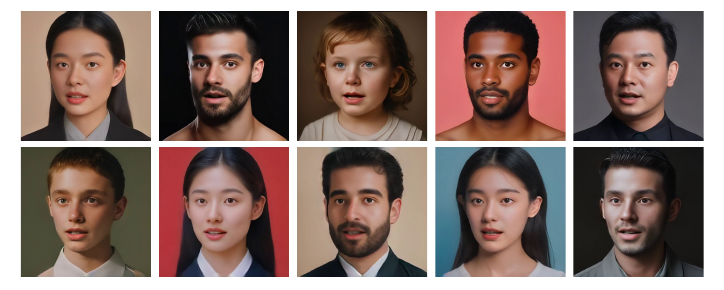}\label{fig:8b}
    \caption{Several samples generated by a single student model.}
    \label{fig:8b}
  \end{subfigure}
    \label{fig:8}
  
  \caption{The exploration to the generalization of our student model.}
\end{figure}

To explore the practical utility of the student model, we further investigated its generalization capability by training it with several different identities. Fig. \ref{fig:8a} reveals that the identity-preserving ability (CSIM) of our student model almost does not drop (0.76 $\rightarrow$ 0.74) when the number of identities increases (1 $\rightarrow$ 20). At the same time, the visual quality (Energy) slightly decreases but still keeps an acceptable level (4.20 $\rightarrow$ 3.70). Visualized results in Fig. \ref{fig:8b} also show that our student model can generalize to multiple speakers while maintaining generation quality, which greatly reduces the deployment cost in practical applications.

\subsection{Ablation Studies} 
\label{sec4d4}
Ablation studies are performed to explore the contribution of each proposed component. For the teacher model, we construct four variants including SSR, early infuse and late infuse in MEM as well as our full model. The quantitative results are shown in Tab. \ref{Tab4}. Obvisouly, the version w/o SSR exhibit degradation in visual quality metrics, while those w/o MEM show decline in motion metrics. It demonstrates that the SSR helps the image clarity and the MEM (infusions) facilitate motion accuracy. 

For the student model, we examined the impact of the delivered features by manipulating three variables: 1) w/o neural keypoints which learns $P_{2D}$ using a network instead of deriving from the teacher model; 2) w/o appearance features which predicts $\mathcal{F}_{2D}$ from the image $I_s$ rather than the teacher's $\mathcal{F}_{3D}$; 3) w/o discriminator which fires the teacher's pretrained discriminator for supervision. The results presented in Tab. \ref{Tab5} clearly reveal that optimal performance is only realized by the full model, verifying the significance of each element we have proposed.

\begin{table}[tb]
  \caption{Ablation studies of teacher model.}
  \label{Tab4}
  \centering
  \begin{tabular}{cccccccccc}
    \toprule
     \multirow{2}{*}{Methods} & \multicolumn{2}{c}{Visual} & \multicolumn{4}{c}{Same-id Motion} & \multicolumn{3}{c}{Cross-id Motion}
     \\
     \cmidrule(l){2-3}
     \cmidrule(l){4-7}
     \cmidrule(l){8-10}
     &Ene$\uparrow$ & Ent/$10^6$$\uparrow$ & CSIM$\uparrow$ & AKD$\downarrow$ & APD$\downarrow$ & AED$\downarrow$ & CSIM$\uparrow$ & APD$\downarrow$ & AED$\downarrow$ \\
    \midrule
    w/o early fusion  & 4.82 & 7.41 & 0.641 & 1.736 & 1.064 & 0.087 & 0.428 & 4.410 & 0.162\\
    w/o late fusion & \textBF{4.89} & \textBF{7.75} & 0.658 & 2.477 & 1.183 & 0.094 & 0.445 & 4.443 & 0.181\\ 
    w/o SSR & 4.79 & 6.65  & \textBF{0.685} & 1.299 & 0.881 & \textBF{0.071} & 0.486 & 4.380 & 0.163\\
    full & 4.86 & 7.59  & 0.682 & \textBF{1.077} & \textBF{0.839} & \textBF{0.071} & \textBF{0.491} & \textBF{3.225} & \textBF{0.151} \\
  \bottomrule
  \end{tabular}
\end{table}

\begin{table}[tb]
  \caption{Ablation studies of student model.
  }
  \label{Tab5}
  \centering
  \begin{tabular}{ccccccc}
    \toprule
     \multirow{2}{*}{Methods} & \multicolumn{2}{c}{Visual} & \multicolumn{3}{c}{Motion}\\
     \cmidrule(l){2-3}
     \cmidrule(l){4-7}
     &Ene$\uparrow$ & Ent/$10^6$$\uparrow$ & CSIM$\uparrow$ & APD$\downarrow$ & AKD$\downarrow $&AED$\downarrow$\\
    \midrule 
    w/o nk  & 4.70 & 3.47  &0.612 & 0.703& 5.004 & 0.148 \\
    w/o app & 4.68 & 3.65  & 0.715 & 0.681 & 0.989 & 0.087\\
    w/o disc & 4.69 & 3.80  & 0.720 & 0.614 & 0.969 & 0.074\\
    full & \textBF{4.78} & \textBF{4.16}& \textBF{0.758} & \textBF{0.399} & \textBF{0.907} &\textBF{0.069}\\
  \bottomrule
  \end{tabular}
\end{table}

\section{Conclusion}
In this work, we propose SuperFace, a superior and pragmatic teacher-student framework for talking head generation. We start by designing an powerful teacher model using the super-resolution training and motion-enhancing mechanism to achieve high quality and robustness. Then we propose a  generic distillation paradigm to distill its knowledge into a computationally efficient architecture. A mask-training-mechanism resolves the issue of semantic decoupling in driving signals, enabling local editing and cross-modal control. Extensive comparisons demonstrate that our SuperFace outperforms state-of-the-art approaches, and the student model inherits high performance from the teacher with exhibits characteristics of low cost. Ablation studies clearly show that our proposed components contribute to model's performance.

\bibliographystyle{splncs04}
\bibliography{main}
\end{document}